\documentclass[10pt,twocolumn,letterpaper]{article}

\usepackage[pagenumbers]{cvpr} %

\usepackage{pifont}

\usepackage{microtype}

\usepackage{placeins}
\usepackage{stfloats}  %

\usepackage{amsmath}

\newcommand{\cmmnt}[1]{\ignorespaces}

\newcommand{\cmark}{{\ding{51}}}
\newcommand{\xmark}{{\ding{55}}}

\newcommand{\sizetwo}[2]{\ensuremath{#1\!\times\!#2\xspace}}
\newcommand{\sizethree}[3]{\ensuremath{#1\mkern-2mu\times\mkern-2mu#2\mkern-2mu\times\mkern-2mu#3\xspace}}

\makeatletter
\DeclareRobustCommand\onedot{\futurelet\@let@token\@onedot}
\def\@onedot{\ifx\@let@token.\else.\null\fi\xspace}

\def\eg{\emph{e.g}\onedot} 
\def\ie{\emph{i.e}\onedot}

\def\etal{\emph{et~al}\onedot}
\makeatother

\newcommand{\muvit}{\mbox{\textsc{MuViT}}\xspace}

\newcommand{\datasynth}{\mbox{\textsc{Synthetic}}\xspace}
\newcommand{\datakidney}{\mbox{\textsc{KPIS}}\xspace}
\newcommand{\datamouse}{\mbox{\textsc{Mouse}}\xspace}

\makeatletter
\newcommand{\saferef}[2]{%
  \@ifundefined{r@#1}{#2}{\ref{#1}}%
}
\makeatother \usepackage{graphicx}
\definecolor{cvprblue}{rgb}{0.21,0.49,0.74}
\usepackage[pagebackref,breaklinks,colorlinks,allcolors=cvprblue]{hyperref}

\title{\muvit: Multi-Resolution Vision Transformers for Learning Across Scales in Microscopy}
\author{Albert Dominguez Mantes \hspace{0.5in} Gioele La Manno$^{\dagger}$\\
Swiss Federal Institute of Technology\\
Lausanne, Switzerland\\
{\tt\small \{albert.dominguezmantes,gioele.lamanno\}@epfl.ch}
\and
Martin Weigert$^{\dagger}$\\
Center for Scalable Data Analytics and Artificial Intelligence (ScaDS.AI)\\Technische Universität Dresden,Germany\\
{\tt\small martin.weigert@tu-dresden.de}
}

\begin{document}
\maketitle

\let\thefootnote\relax\footnotetext{
\begin{tabular}{@{}l@{}}
$^{\dagger}$Co-corresponding authors. \\
Code: \href{https://github.com/weigertlab/muvit}{\color{cvprblue}github.com/weigertlab/muvit}
\end{tabular}
}

\begin{abstract}
    Modern microscopy routinely produces gigapixel images that contain structures across multiple spatial scales, from fine cellular morphology to broader tissue organization. Many analysis tasks require combining these scales, yet most vision models operate at a single resolution or derive multi-scale features from one view, limiting their ability to exploit the inherently multi-resolution nature of microscopy data.
    We introduce \muvit, a transformer architecture built to fuse true multi-resolution observations from the same underlying image. \muvit embeds all patches into a shared world-coordinate system and extends rotary positional embeddings to these coordinates, enabling attention to integrate wide-field context with high-resolution detail within a single encoder.
    Across synthetic benchmarks, kidney histopathology, and high-resolution mouse-brain microscopy, \muvit delivers consistent improvements over strong ViT and CNN baselines. Multi-resolution MAE pretraining further produces scale-consistent representations that enhance downstream tasks. These results demonstrate that explicit world-coordinate modelling provides a simple yet powerful mechanism for leveraging multi-resolution information in large-scale microscopy analysis.
\end{abstract}
\section{Introduction}
\label{sec:intro}

Modern microscopy techniques generate images at unprecedented scale and complexity, with modalities such as light-sheet fluorescence microscopy~\cite{yao2023high,yang2022daxi,glaser2025expansion}, electron microscopy~\cite{xu2021open,muller2024modular}, and digital pathology~\cite{chen_scaling_2022,xu2024whole} routinely producing gigapixel-scale images exceeding $\sizetwo{50K}{50K}$ pixels. 
These images capture biological structures with hierarchical organization across spatial scales, from individual cells to tissue architecture to anatomical regions.
A central challenge in the computational analysis of such images is that many tasks require information across these scales simultaneously. For instance, accurate semantic segmentation of local cellular structures often depends on broader anatomical context, \eg  knowing whether a cell is part of a particular tissue region can be essential for correct classification. In the last decade, deep learning approaches based on CNNs~\cite{ronneberger2015,stringer2021cellpose,schmidt2018cell} or vision transformers~\cite{dosovitskiy_vit_2021,archit2025segment} have emerged as the standard for segmentation and detection in microscopy. 
These approaches, however, typically operate via tiled prediction on single-resolution crops (\eg of size 512$\times$512 pixels) for memory reasons, which restricts the available context to the processed tile. As a consequence, they must trade off field of view against spatial resolution, limiting their ability to access fine detail and global context simultaneously. 

\begin{figure*}[t]
    \centering
    \includegraphics[width=1.\linewidth]{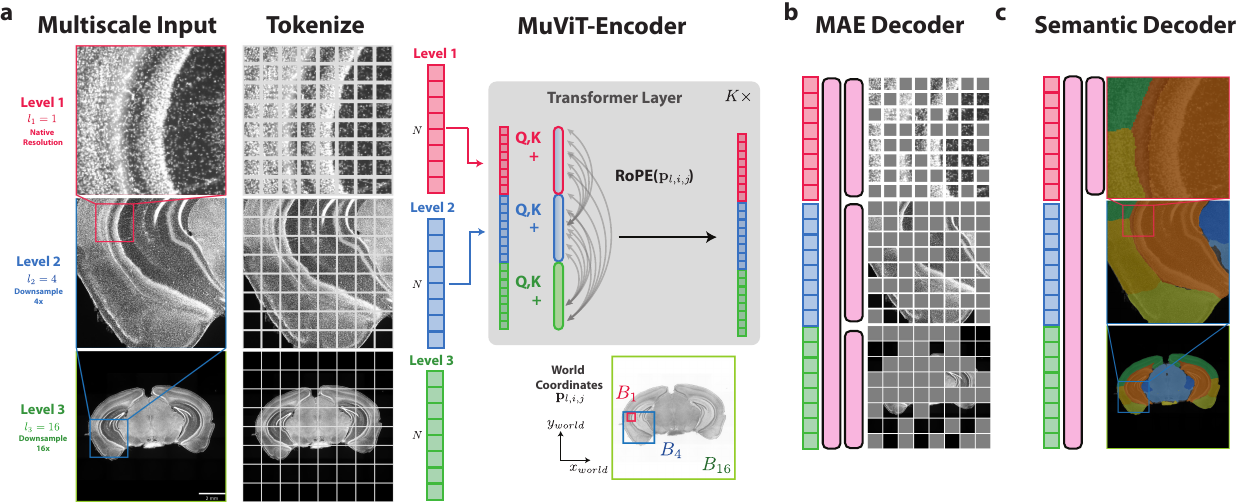}
    \caption{Overview of the \muvit architecture. \textbf{a)} \muvit processes multiple crops of the same scene at different physical resolutions and encodes them jointly using a shared transformer with resolution-specific RoPE position encodings. As a decoder we use \textbf{b)} a lightweight masked autoencoding decoder that reconstructs the masked patches for each resolution level, or \textbf{c)} a semantic segmentation decoder that predicts the class of each pixel at the target resolution (\eg level 1).}
    \label{fig:overview}
\end{figure*}

In this work, we propose \muvit (\textbf{Mu}lti-Resolution \textbf{Vi}sion \textbf{T}ransformer), an architecture that jointly processes several crops of the same image captured at different physical resolutions. Throughout this work, we use the term \emph{resolution} to denote the physical scale of an observation (coarse versus fine), rather than differences in input pixel dimensions.
Although some transformers are called resolution-independent, this usually refers to handling different input sizes rather than modeling multiple physical scales~\cite{fan2024vitar}.
Likewise, hierarchical architectures that construct feature pyramids from a single input~\cite{liu_mmvit_2023} and models that derive multi-scale features only through internal downsampling, such as U-Net~\cite{ronneberger2015} or pyramidal vision transformers~\cite{wang2021pyramid}, do not access true multi-resolution information.

Our design builds on the idea that different spatial scales can be treated as complementary input ``modalities'' similar  to multi-modal encoders such as MultiMAE~\cite{bachmann_multimae_2022}, as long as they are linked through a shared geometric reference frame. \muvit accepts spatially related crops that share pixel size but differ in field of view.
To ensure geometric consistency across these crops, we incorporate common world-coordinates (which we define as the pixel coordinate system of the highest-resolution input) per token using Rotary Position Embeddings (RoPE)~\cite{su_roformer_2022} which applies these coordinates directly within the internal query and key projections. This allows \muvit to perform explicit cross-resolution attention across different physical scales.
Our contributions are: 
\emph{i)} We propose \muvit, a multi-resolution Vision Transformer architecture that jointly processes true multi-resolution observations within a single encoder, in contrast to hierarchical methods that construct feature pyramids from a single-resolution input. 
\emph{ii)} We incorporate absolute world coordinates into attention via RoPE, enabling cross-resolution interactions without crop alignment, showing that accurate coordinate relations are crucial for effective downstream performance.
\emph{iii)} We extend MAE pretraining to the multi-resolution setting and show that adding resolution levels yields increasingly informative representations. Multi-resolution MAE also accelerates downstream training, with segmentation models converging within only a few epochs.
\emph{iv)} We demonstrate strong downstream performance of the resulting \muvit backbone on both synthetic and large-scale microscopy tasks, achieving strong performance on the \datakidney benchmark~\cite{tang2024} and substantial gains over single-resolution encoders on mouse brain anatomy segmentation.

\section{Related work}
\label{sec:related}

\paragraph{Multi-scale learning with transformers.}
Existing transformer-based methods incorporate multi-scale information in fundamentally different ways. 
Some approaches construct all scales internally from a single input resolution, such as hierarchical or pyramidal models like Swin~\cite{liu2021swin}, PVT~\cite{wang2021pyramid}, and HIPT~\cite{chen_scaling_2022}. Others train on multiple input resolutions for robustness but still infer at a single resolution, as in ResFormer~\cite{tian_resformer_2023}. Parallel multi-path models process artificially created scale variants through separate branches, including MMViT~\cite{liu_mmvit_2023}, CrossViT~\cite{chen_crossvit_2021}, MPViT~\cite{lee_mpvit_nodate}, and MUSIQ~\cite{ke_musiq_2021}. 
All of these derive multi-scale features from either a single crop or independent inputs and lack explicit geometric correspondence across scale levels, as is also the case for global-local fusion approaches~\cite{chen2019collaborative}.
In contrast, \muvit integrates multiple physically distinct resolutions extracted from a larger image within a unified encoder, using world-coordinate embeddings to align their geometry and enable cross-scale attention.

\paragraph{Masked autoencoder pre-training.}
Masked autoencoders (MAEs)~\cite{he_masked_2022} pre-train a ViT encoder by reconstructing heavily masked inputs using a lightweight decoder and have proven effective across a wide range of vision tasks. 
MultiMAE~\cite{bachmann_multimae_2022} extends this paradigm to multi-modal inputs by treating different modalities (e.g., RGB and depth) as separate token types with cross-modality reconstruction. Nevertheless, it assumes trivial spatial alignment at a single resolution.
Motivated by this, \muvit treats different spatial scales as distinct but geometrically related input types and uses Dirichlet-weighted masking across resolution levels to promote cross-scale learning within a shared coordinate frame. 
Scale-MAE~\cite{reed_scale-mae_2023} addresses heterogeneous image resolutions in geospatial datasets by encoding scale (via GSD) on single-resolution inputs, but does not perform joint encoding or geometric alignment across multiple FOVs.
Cross-Scale MAE~\cite{tang_cross-scale_nodate} enforces consistency across scale-augmented views of the same image.
Kraus~\etal~\cite{kraus_masked_2024} adapt MAEs to fluorescence microscopy, but only for single-resolution inputs.
Unlike these approaches, \muvit performs joint masked reconstruction across \emph{physically distinct} resolution levels of the same scene using world-coordinate RoPE to maintain geometric alignment between scales.

\begin{table}[t]
    \centering
    \resizebox{\columnwidth}{!}{%
    \begin{tabular}{lcccc}
        \toprule
        \textbf{Method} 
        & \textbf{Hierarchical} 
        & \textbf{Physical} 
        & \textbf{Joint} 
        & \textbf{World Coordinate} \\
        & \textbf{Stages} 
        & \textbf{Multi-Res Input} 
        & \textbf{Attention} 
        & \textbf{Alignment} \\
        \midrule
        MultiMAE   
        & \xmark & \xmark & \cmark & \xmark \\
        Scale-MAE 
        & \xmark & \xmark & \xmark & \xmark \\
        PVT / Swin 
        & \cmark & \xmark & \xmark & \xmark \\
        HIPT       
        & \cmark & \cmark & \xmark & \xmark \\

        \textbf{\muvit (Ours)} 
        & \textbf{\xmark} & \textbf{\cmark} & \textbf{\cmark} & \textbf{\cmark} \\
        \bottomrule
    \end{tabular}%
    }
    \caption{Conceptual comparison of relevant related work.}
    \label{tab:novelty}
\end{table} 

\paragraph{Rotary positional embeddings in vision.}
While RoPE~\cite{su_roformer_2022} has become the default positional encoding for large language models, most ViT architectures still use alternative schemes such as fixed Fourier features~\cite{dosovitskiy_vit_2021,kirillov2023segment}. Early adaptations of RoPE to vision include EVA-02~\cite{fang_eva-02_2024}, which applies RoPE independently to each spatial axis, and RoPE-Mixed~\cite{heo_rotary_2024}, which learns axis-frequency mixtures to better capture diagonal structure in high-resolution images. Further extensions such as ComRoPE~\cite{yu_comrope_nodate} and LieRE~\cite{ostmeier_liere_2025} propose more general parametrizations through trainable matrices and Lie group formulations. 
In contrast to these design-oriented extensions, \muvit uses standard axis-wise RoPE in a new setting: the rotation angles are derived from each tokens world-coordinates, enabling geometrically consistent attention across resolution levels of the same scene.

\section{Methods}
\label{sec:methods}

\subsection{Input and spatial representation}

\muvit receives multi-resolution image crops and their spatial extents as input. The input consists of a tuple $(\mathbf{X}, \mathcal{B})$ where $\mathbf{X} \in \mathbb{R}^{L \times C \times H \times W}$ contains the image crops at $L$ resolution levels, $\mathcal{B} \in \mathbb{R}^{L \times 2 \times 2}$ defines the bounding box of each crop, $C$ is the number of channels, and $H, W$ are the spatial dimensions in pixels. $\mathcal{L} = [l_1, l_2, \ldots, l_L]$ with $l_i \in \mathbb{N}$ specifies the downsampling factor for each resolution level.
We use $l=1$ for the finest resolution; $l>1$ denotes $l\times$ downsampling, so an $H\times W$ crop at level $l$ spans the same scene region as an $(lH)\times(lW)$ crop at level 1.

 $H$ and $W$ remain constant across levels in pixel space, so lower resolution levels capture larger spatial context at the cost of fine-grained detail. We denote by \muvit{}$_{[l_1, l_2, \ldots]}$ an encoder using resolution levels $l_1, l_2, \ldots$ so that a \muvit{}$_{[1,8]}$ encoder processes the original resolution and an 8$\times$ downsampled version.
The bounding box $\mathcal{B}_l = ((\text{y}^\text{min}_{l}, \text{x}^\text{min}_l), (\text{y}^\text{max}_l, \text{x}^\text{max}_l))$ defines the field of view for each crop in world coordinates, which we define as the pixel coordinate system of the highest-resolution input (\ie level $l=1$).

\paragraph{Bounding box generation.} Bounding boxes for all  resolution levels are derived from the same world-coordinate extent, ensuring geometric alignment. During training, we sample nested crops at arbitrary positions and provide the model with their true bounding box coordinates for accurate spatial encoding. To test the importance of spatial information, we also evaluate a baseline where each level is assigned a fixed, centered bounding box scaled by its downsampling factor (\textit{naive bbox}), independent of the actual crop location. As shown in~\cref{sec:experiments} such incorrect bounding box information leads to a substantial performance drop, confirming that accurate spatial information crucial for effective cross-scale learning.

For each resolution level $l$, non-overlapping patches (tokens) of size $P \times P$ (we use $P=8$) are extracted and projected to a $D$-dimensional embedding space through level-specific linear layers followed by layer normalization, denoted as $\text{PE}_{l}$. To distinguish tokens from different resolution levels, learnable level embeddings $\mathbf{e}_{l}$ are added: $\mathbf{z}_l = \text{PE}_{l}(\mathbf{X}_l) + \mathbf{e}_{l}$, where $\mathbf{z}_l \in \mathbb{R}^{N \times D}$ with $N = (H/P) \times (W/P)$ patches per level. In addition to the embedded tokens, we compute spatial coordinates for each patch from the bounding boxes. For a patch at position $(i,j)$ in level $l$, its center coordinate in world space is: $\mathbf{p}_{l,i,j} = \mathcal{B}_l^{\text{min}} + \left(\tfrac{i}{h-1}, \tfrac{j}{w-1}\right) \odot \left(\mathcal{B}_l^{\text{max}} - \mathcal{B}_l^{\text{min}}\right)$
where $h = H/P$ and $w = W/P$, and $\odot$ denotes element-wise multiplication. These coordinates encode the absolute spatial position of each patch in a common world coordinate system, enabling consistent positional encoding across resolution levels. The concatenated tokens $\mathbf{z} = [\mathbf{z}_1; \ldots; \mathbf{z}_L] \in \mathbb{R}^{(L \cdot N) \times D}$ and their coordinates $\mathbf{p} = [\mathbf{p}_1; \ldots; \mathbf{p}_L] \in \mathbb{R}^{(L \cdot N) \times 2}$ are processed by the \muvit encoder.

\subsection{\muvit encoder}

The \muvit encoder consists of $K$ stacked transformer layers (we use $K=12$ for all experiments) that process the full sequence of multi-resolution tokens jointly through a unified architecture. Unlike standard ViT architectures that apply fixed Fourier-based positional encodings to input tokens, \muvit uses 2D axial Rotary Position Embeddings (RoPE)~\cite{su_roformer_2022,fang_eva-02_2024} computed from world coordinates $\mathbf{p}$. For each spatial dimension (y and x), the head dimension $d_h = D/H$ is split into per-axis allocations $d_y$ and $d_x$ such that $d_y + d_x \leq d_h$. Each transformer layer has independent learnable RoPE parameters. For each coordinate axis $a \in \{y, x\}$ and patch position $\mathbf{p}_{l,i,j}^{(a)}$, the rotation frequencies are:
\begin{equation}
    \theta_k^{(a)} = \mathbf{p}_{l,i,j}^{(a)} / b^{2k/d_a}, \quad k = 0, \ldots, d_a/2 - 1
\end{equation}
where the frequencies $b^{2k/d_a}$ are a learnable parameter (initialized to $b=10000$) allowing each layer to adapt its positional encoding. 
Crucially, computing the frequencies from the world coordinates $\mathbf{p}$ ensures that patches representing the same spatial location receive identical positional encodings regardless of their resolution level $l$ thus enabling effective cross-scale information flow. This differs from learned positional embeddings, which would treat the same spatial location differently at each resolution. 
As we show in \cref{sec:experiments}, accurate world-coordinates are essential: replacing them with naive centered (\ie inconsistent) coordinates leads to a collapse in performance, even though the architecture and inputs remain unchanged. Nevertheless, trained \muvit models exhibit a certain degree of robustness to world coordinate misalignment (\cref{sec:experiments}).

Each transformer layer follows a standard structure with residual connections:
\begin{align}
    \mathbf{x}^{(i)} &\leftarrow \mathbf{x}^{(i-1)} + \text{SelfAttn}(\text{LN}(\mathbf{x}^{(i-1)}), \mathbf{p}) \\
    \mathbf{x}^{(i)} &\leftarrow \mathbf{x}^{(i)} + \text{FFN}(\text{LN}(\mathbf{x}^{(i)}))
\end{align}
where LN denotes layer normalization and FFN is a feed-forward network with GELU activation.  Given the input image crops and bounding boxes $(\mathbf{X}, \mathcal{B})$, the encoder produces encoded features $\mathbf{h} \in \mathbb{R}^{(L \cdot N) \times D}$ that capture multi-resolution representations with explicit world-coordinate relationships. We denote by \muvit{}$_{[l_1, l_2, \ldots]}$ an encoder using levels $l_1, l_2, \ldots$, and by \muvit{}$_{[\ldots]}$+decoder the same encoder paired with a specific decoder (\eg segmentation). The full encoder  is lightweight containing $\sim$25M parameters.

\subsection{\muvit-MAE: Self-supervised pre-training}

We adapt masked autoencoding to the multi-resolution setting.  During training, a random subset of tokens is masked with a high masking ratio $\rho$ (we use $\rho=0.75$). To encourage the model to learn diverse cross-scale configurations, the proportions of visible tokens per level $(\lambda_1,\ldots,\lambda_L)$ are sampled from a Dirichlet distribution $\text{Dir}(\alpha)$ with $\alpha=0.5$~\cite{bachmann_multimae_2022}. The encoder processes only the visible tokens and their world-coordinates:
\begin{equation}
    \mathbf{h}_{\text{vis}} = \text{Encoder}(\mathbf{z}_{\text{vis}}, \mathbf{p}_{\text{vis}}).
\end{equation}
To reconstruct the masked tokens, we use $L$ lightweight decoders, one per resolution level, each consisting of 2 transformer layers. Learnable mask tokens are inserted at masked positions, and the first decoder layer incorporates cross-attention to the full set of visible encoder outputs with coordinate-based RoPE, enabling information flow across scales. Each decoder reconstructs the masked patches for its corresponding resolution level through a linear projection to pixel space. As loss we use mean squared error on masked patches only, averaged across all resolution levels.

\subsection{\muvit-seg: Semantic segmentation}

For semantic segmentation tasks, we use the pre-trained \muvit encoder to extract multi-resolution features and add task-specific decoders that we fine-tune or keep frozen during training. We explore two decoder designs: the first, inspired by UNETR~\cite{hatamizadeh2022}, progressively upsamples features through a series of convolutional blocks while incorporating skip connections from intermediate encoder layers. For our 12-layer encoder with patch size 8, skip connections are taken from layers [0, 5, 11], corresponding to three decoder stages. At each decoder stage, features from the corresponding encoder layer are concatenated with the upsampled features, combining fine-grained multi-scale information from the encoder with the coarse semantic features from the decoder. The second design follows Mask2Former~\cite{cheng2022masked} and uses learnable mask queries that cross-attend to the multi-resolution encoder features. For both architectures, we only use the finest decoder level as the final prediction while using the following loss function:
\begin{equation}
    \mathcal{L} = \lambda_{\text{CE}} \cdot \mathcal{L}_{\text{CE}}(\tilde{y}, y) + \lambda_{\text{Dice}} \cdot \mathcal{L}_{\text{Dice}}(\tilde{y}, y)
\end{equation}
with $\tilde{y}$ the predicted segmentation and $y$ the ground truth (we use $\lambda_{\text{CE}}=\lambda_{\text{Dice}}=1.0$). The loss is computed only at the finest resolution level ($l=1$), where final segmentation is produced.
\section{Experiments}
\label{sec:experiments}

We evaluate \muvit on three datasets that each address different aspects: a synthetic dataset designed to establish a controlled setting for validation of cross-scale behavior, a large-scale anatomical segmentation task requiring strong global/local integration, and a challenging pathology benchmark dataset that allows to externally quantify the performance of the model.

\subsection{Datasets}

\textbf{Synthetic dataset (\datasynth).} We generate a synthetic dataset with controlled multi-resolution structure to validate the architecture. The dataset consists of 1638 training images, 220 validation images, and 190 test images of size 2048$\times$2048 pixels (single channel) with procedurally generated concentric ring patterns using Perlin noise. The task is to segment two object classes: outer cells and inner cells. This task requires multi-resolution processing as the pixel class depends on its position relative to the global ring structure, which is only visible at coarser resolutions. This controlled setting allows us to focus on the impact of multi-resolution processing and to validate the benefits of using scale-consistent positional encodings.

\textbf{Mouse brain anatomy dataset (\datamouse).} The mouse brain anatomy dataset~\cite{staeger2020} consists of DAPI-stained microscopy slices from 12 volumetric mouse brains with anatomical region annotations for multi-class semantic segmentation. The dataset contains 618 images of approximately \sizetwo{13k}{9k} pixels, which we split into 563 training images and 55 test images. The task is to segment 11 distinct anatomical brain regions (\ie 12 classes total including the background class). Multi-resolution processing is critical because correctly classifying a pixel requires understanding which global anatomical region it belongs to, while fine-resolution information is necessary for precise boundary delineation. We use a specimen-based split with specimen BM15 being held out as the test set. 

\textbf{Kidney pathology segmentation (\datakidney).} The \datakidney dataset~\cite{tang2024} consists of RGB histopathology whole-slide images (WSIs) of mouse kidney tissue of four different groups of mice models for glomeruli segmentation. The dataset contains 30 training images, 8 validation images, and 12 test images with sizes ranging from \sizetwo{16k-71k}{14k-81k} pixels. The task to solve consists of identifying pixels of glomerular structures (binary semantic segmentation) across the whole WSIs. We hypothesized that multi-resolution processing is beneficial because glomeruli can be larger than the receptive field of commonly used architectures. They may also be located in specific anatomical regions of the tissue, thus requiring both local texture information for identification and global tissue context for localization. As the WSIs are obtained from four different groups of mice models representing different conditions and stages of chronic kidney disease, we also use this dataset to assess the representation power of MAE pretrained models in a linear probing experiment.

\textbf{Implementation details.} All models are implemented in PyTorch and trained using PyTorch Lightning. Full training details, hyperparameters, and optimization settings are provided in Supplementary Section \saferef{sec:suppl_training}{A}. During training, we sample nested multi-resolution crops by randomly generating coordinates at the finest scale and then successively sample at coarser scales while ensuring the crop of resolution level $l_i$ is always contained in the crop of level $l_{i+1}$. Due to the large size of the images, we use Zarr pyramids to store the datasets. Zarr's chunked storage standard allows to load only the sampled positions at multiple scales, enabling efficient memory usage and fast data loading during training and inference.

\begin{figure*}[t]
    \centering
    \includegraphics[width=\linewidth]{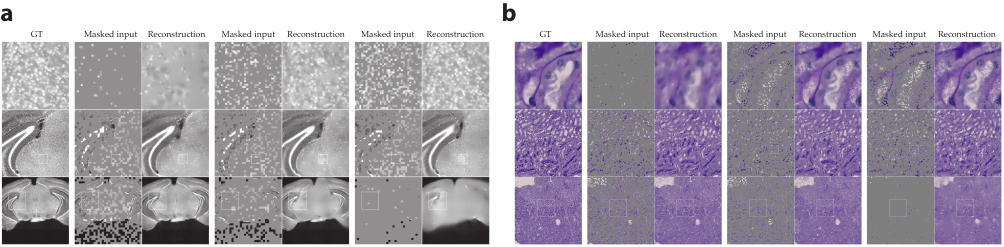}
    \caption{\muvit-MAE reconstruction results on \textbf{a)} \datamouse and \textbf{b)} \datakidney. Columns show masked input and reconstruction at multiple resolution levels ($l_1=1$ top, $l_2=8$ middle, $l_3=32$ bottom). The same overall masking ratio ($\rho=0.75$) is used across examples, but each column demonstrates different Dirichlet-sampled per-level masking distributions.}
    \label{fig:mae}
\end{figure*}
\subsection{Evaluation metrics}

We evaluate semantic segmentation performance using the Dice coefficient. The Dice coefficient for class $c$ is defined as $\text{DSC}_c = \frac{2 |P_c \cap G_c|}{|P_c| + |G_c|}$

where $P_c$ and $G_c$ are the predicted and ground truth pixels for class $c$, respectively. For the multi-class case (\datamouse dataset), we report both per-class DICE scores and their macro average (mDSC).

\subsection{MAE pretraining}

We pretrain \muvit{} encoders using MAE on all datasets, including \datasynth, \datamouse and \datakidney. As shown in \Cref{fig:mae}, MAE models successfully reconstruct masked patches across multiple resolution levels. The Dirichlet-sampled masking strategy described in \Cref{sec:methods} encourages the model to learn cross-scale relationships by varying the relative masking ratio per level while maintaining an overall masking ratio ($\rho=0.75$). \Cref{fig:mae} depicts results for different per-level masking distributions, with the model consistently reconstructing fine details at high resolution while maintaining global coherence across scales.

\begin{table}[b]
  \caption{MAE ablation study on \datamouse and \datakidney. Reconstruction mean squared error is reported on 100 nested multi-resolution samples per image (test set). \muvit$_{[l]}$ denotes an MAE model that only receives the evaluated level. }
  \label{tab:mae_ablation}
  \centering
  \resizebox{\columnwidth}{!}{%
  \begin{tabular}{@{}llcccc@{}}
    \toprule
    Dataset & Method & Level 1 & Level 8 & Level 32 & Level 64 \\
    \midrule
    \datamouse & \muvit$_{[l]}$ & $5.192\times 10^{-3}$ & $1.628\times 10^{-2}$ & $1.671 \times 10^{-2}$ & -\\
               & \muvit$_{[1,8,32]}$ {\small (naive)} & $\mathbf{4.848 \times 10^{-3}}$ & $\mathbf{1.589 \times 10^{-2}}$ & $\mathbf{1.373 \times 10^{-2}}$ & -\\
               & \muvit$_{[1,8,32]}$ & $8.734 \times 10^{-3}$ & $2.274 \times 10^{-2}$ & $1.464 \times 10^{-2}$ & -\\
    \midrule
    \datakidney & \muvit$_{[l]}$ & $4.352 \times 10^{-3}$ & $1.122 \times 10^{-2}$ & $1.013\times 10^{-2}$ & $1.021\times 10^{-2}$\\
                &  \muvit$_{[1,8,32,64]}$ {\small (naive)} & $3.410 \times 10^{-3}$ & $9.998 \times 10^{-3}$ & $8.816 \times 10^{-3}$ & $7.302\times 10^{-3}$\\
                &  \muvit$_{[1,8,32,64]}$ & $\mathbf{3.268 \times 10^{-3}}$ & $\mathbf{9.387 \times10^{-3}}$ & $\mathbf{7.941 \times10^{-3}}$ & $\mathbf{6.913 \times10^{-3}}$\\
    \bottomrule
  \end{tabular}%
  }
\end{table}
 \Cref{tab:mae_ablation} shows the reconstruction error for \datamouse and \datakidney comparing reconstruction error without (\textit{naive}) and with the default scale-consistent bounding box coordinates across different resolution levels. The results demonstrate that accurate coordinate alignment is crucial for reconstruction at coarser scales (\eg levels $l=8/32/64$), where knowing the high resolution view's location relative to the coarse scales enables improved reconstruction of the latter ones. At the finest resolution ($l=1$), the model achieves a low reconstruction error even with naive coordinates, suggesting that local texture patterns provide sufficient cues without global context. This suggests coarse-scale tokens rely on world-coordinate alignment to access complementary high-resolution information, whereas fine-scale tokens can be reconstructed from local appearance alone.

To evaluate how MAE pretraining provides high-quality initial representations, \Cref{tab:training_progress} reports validation Dice scores during fine-tuning on \datamouse. Models initialized with MAE-pretrained \muvit encoders converge dramatically faster than all baselines: \muvit$_{[1,8,32]}$+Mask2Former reaches m\text{DSC}=0.843 after only 10 epochs and already exceeds the final performance of every single-resolution model (and surpasses 0.88 by epoch 25). In contrast, conventional architectures trained on large $\sizetwo{1024}{1024}$ inputs show slow and unstable early training, with most models below 0.30 Dice at epoch 10. This demonstrates that MAE pretraining yields high-quality initial features across scales that substantially accelerate optimization and improve sample efficiency.
\begin{table}[b]
\caption{\datamouse training progress results. Validation mDSC (11 classes) is reported at different epochs during training.}
\resizebox{\columnwidth}{!}{%
\begin{tabular}{lccccc}
\toprule
Method & Input Size & \multicolumn{4}{c}{Epoch} \\
\cmidrule(lr){3-6}
& & E1 & E10 & E25 & E50 \\
\midrule
U-Net & $\sizetwo{1024}{1024}$ & 0.086 & 0.110 & 0.288 & 0.412 \\
DeepLabV3 & $\sizetwo{1024}{1024}$ & 0.234 & 0.582 & 0.634 & 0.704 \\
SegFormer & $\sizetwo{1024}{1024}$ & 0.045 & 0.159 & 0.266 & 0.332 \\
SwinUNETRV2 & $\sizetwo{1024}{1024}$ & 0.049 & 0.232 & 0.305 & 0.388 \\
\midrule
\muvit$_{[1]}$ {\small +Mask2Former} & $\sizetwo{256}{256}$ & 0.033 & 0.142 & 0.266 & 0.310 \\
\muvit$_{[1]}$ {\small +UNETR} & $\sizetwo{256}{256}$ & 0.105 & 0.086 & 0.261 & 0.281 \\
\muvit$_{[1,8,32]}$ {\small +Mask2Former} & $\sizethree{3}{256}{256}$ & \textbf{0.619} & \textbf{0.843} & \textbf{0.880} & \textbf{0.878} \\
\muvit$_{[1,8,32]}$ {\small +UNETR} & $\sizethree{3}{256}{256}$ & 0.297 & 0.748 & 0.852 & 0.864 \\
\bottomrule
\end{tabular}
}%
\label{tab:training_progress}
\end{table}

\subsection{Semantic segmentation on \datasynth}

\begin{figure*}[t]
    \centering
    \includegraphics[width=\linewidth]{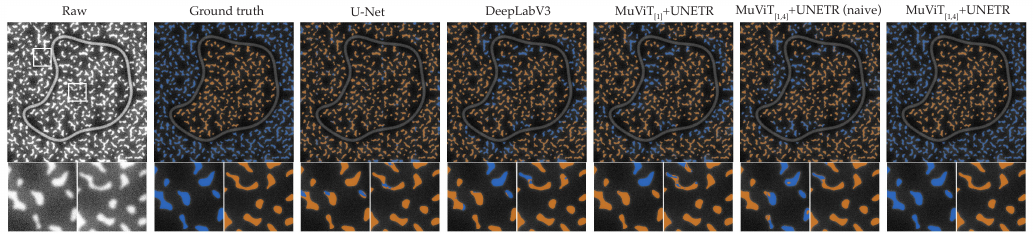}
    \caption{Semantic Segmentation results on \datasynth. Columns show: input image, ground truth, and semantic predictions for different architectures (U-Net, DeepLabV3, \muvit$_{[1]}$,  \muvit$_{[1,4]}$+UNETR (naive, \ie without cross-scale consistent positional encodings), and \muvit$_{[1,4]}$+UNETR). Top: full image ($\sizetwo{2k}{2k}$ px). Bottom: magnified regions (white boxes) showing detail comparison.}
     \label{fig:seg_synthetic}
  \end{figure*}

To assess the role of multi-resolution processing in a controlled setting, we first evaluate on the \datasynth{} dataset using two levels ([1,4]). During training we sample nested crops at different resolution levels at random global positions. Level~1 crops are always contained within the level~4 crop but not centered in order to force the model to rely on diverse spatial relationships.

\Cref{tab:synthetic} shows that \muvit{}$_{[1,4]}$+UNETR accurately solves the task ($mDSC=0.9538$), whereas all single-resolution baselines can recognize the objects but fail to classify them into the correct class ($mDSC\approx$0.50). Crucially, training the same \muvit architecture with incorrect bounding box coordinates (\textit{naive}) causes the performance to collapse to baseline levels ($mDSC=0.3864$). This demonstrates that the scale-consistent positional encoding is an essential factor: without correct spatial alignment the model is not able to integrate cross-level information.

\begin{table}[t]
  \caption{Semantic segmentation results on \datasynth. Reported are Dice scores for each class (outer, inner) and mDSC. All baselines operate at the highest resolution (corresponding to $l=1$).}
  \label{tab:synthetic}
  \centering
  \begin{tabular}{@{}lrrr@{}}
    \toprule
    Method & outer & inner & mDSC \\
    \midrule
    U-Net & 0.5433 & 0.2600 & 0.4016 \\
    DeepLabV3 & 0.5473 & 0.4317 & 0.4895 \\
    \muvit{}$_{[1]}${\small +UNETR} & 0.5391 & 0.4615 & 0.5003 \\
    \muvit{}$_{[1,4]}${\small +UNETR} {\small (naive)} & 0.4662 & 0.3065 & 0.3864 \\
    \muvit{}$_{[1,4]}${\small +UNETR} & \textbf{0.9499} & \textbf{0.9577} & \textbf{0.9538} \\
    \bottomrule
  \end{tabular}
\end{table}
 
\subsection{Semantic segmentation on \datamouse}

\begin{figure*}[htbp!]
    \centering
    \includegraphics[width=\linewidth]{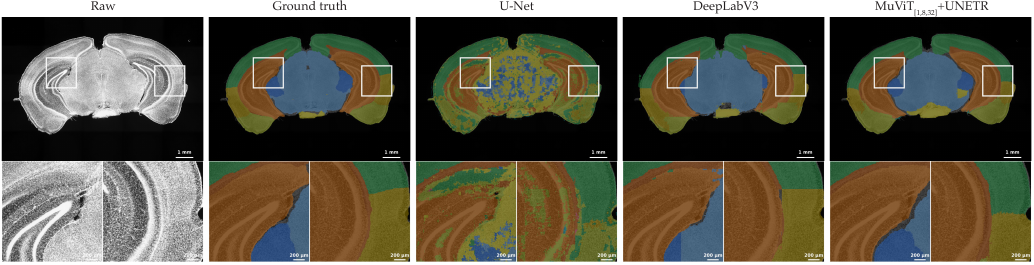}
    \caption{Semantic segmentation results on \datamouse. Columns show: input image, ground truth, U-Net, DeepLabV3, and \muvit{}$_{[1,8,32]}$+UNETR semantic predictions. Top: full image ($\sizetwo{13k}{10k}$ px). Bottom: magnified regions (white boxes) showing detail comparison.}
     \label{fig:seg_mouse}
\end{figure*}

We train several baseline models of different complexities (\saferef{tab:brain_params}{}) along with \muvit on \datamouse. \Cref{tab:brain} shows results on the mouse brain anatomy segmentation task. All methods use tiled prediction at the highest resolution scale, without overlap between tiles. The tile size matches the spatial input that each method was trained on. \muvit{}$_{[1,8,32]}$+Mask2Former achieves a $mDSC=0.901$, substantially outperforming all baselines. The same architecture achieves the highest Dice coefficient in all classes but one (\textit{Caudoputamen}), for which \muvit{}$_{[1,8,32]}$+UNETR performs best. The improvement of the $\muvit{}_{[1,8,32]}$ variants over the rest of the methods is particularly pronounced when baselines are trained with smaller input sizes (\sizetwo{256}{256}), highlighting again that necessity of sufficient spatial context for the task.

\begin{table*}[b]
    \centering
    \caption{Semantic segmentation results on \datamouse (test set). Shown are DSC for each anatomical class as well as averaged (mDSC). All baselines operate at the highest resolution (corresponding to $l=1$).}
    \resizebox{\textwidth}{!}{%
    \begin{tabular}{lccccccccccccc}
    \toprule
    Method & Input Size & \multicolumn{11}{c}{Semantic class} & mDSC \\
    \cmidrule(lr){3-14}
    & & \footnotesize{Hippocampus} & \footnotesize{Thalamus} & \footnotesize{Hypothalamus} & \footnotesize{Septal complex} & \footnotesize{Caudoputamen} & \footnotesize{Dorsal cortex} & \footnotesize{Ventral cortex} & \footnotesize{Lateral cortex} & \footnotesize{Corpus callosum} & \footnotesize{Ventricular system} & \footnotesize{Midbrain} & \\
    \midrule
    U-Net~\cite{ronneberger2015} & $1024 \times 1024$ & 0.777 & 0.489 & 0.515 & 0.112 & 0.802 & 0.780 & 0.546 & 0.216 & 0.533 & 0.480 & 0.149 & 0.491 \\
     & $512 \times 512$ & 0.651 & 0.310 & 0.409 & 0.000 & 0.742 & 0.677 & 0.378 & 0.189 & 0.447 & 0.268 & 0.000 & 0.370 \\
     & $256 \times 256$  & 0.027 & 0.000 & 0.121 & 0.000 & 0.274 & 0.234 & 0.141 & 0.000 & 0.135 & 0.000 & 0.000 & 0.085 \\
    \midrule
    DeepLabV3~\cite{chen2018encoder} & $1024 \times 1024$ & 0.941 & 0.863 & 0.809 & 0.734 & 0.908 & 0.943 & 0.874 & 0.756 & 0.782 & 0.857 & 0.807 & 0.843 \\
     & $512 \times 512$ & 0.894 & 0.675 & 0.660 & 0.278 & 0.851 & 0.878 & 0.721 & 0.472 & 0.670 & 0.747 & 0.585 & 0.676 \\
     & $256 \times 256$  & 0.799 & 0.509 & 0.543 & 0.092 & 0.782 & 0.761 & 0.540 & 0.289 & 0.507 & 0.545 & 0.349 & 0.520 \\
    \midrule
    SwinUNETRV2~\cite{he2023swinunetr} & $1024 \times 1024$ & 0.850 & 0.510 & 0.448 & 0.020 & 0.756 & 0.882 & 0.706 & 0.226 & 0.621 & 0.699 & 0.000 & 0.520 \\
     & $512 \times 512$ & 0.810 & 0.342 & 0.430 & 0.005 & 0.753 & 0.829 & 0.641 & 0.147 & 0.472 & 0.557 & 0.000 & 0.453 \\
     & $256 \times 256$  & 0.707 & 0.283 & 0.310 & 0.000 & 0.691 & 0.700 & 0.454 & 0.000 & 0.307 & 0.216 & 0.000 & 0.333 \\
    \midrule
    \muvit$_{[1]}$ {\small +Mask2Former} & $256 \times 256$  & 0.735 & 0.393 & 0.435 & 0.000 & 0.766 & 0.778 & 0.543 & 0.227 & 0.012 & 0.249 & 0.157 & 0.391 \\
    \muvit$_{[1]}$ {\small +UNETR} & $256 \times 256$  & 0.779 & 0.410 & 0.311 & 0.000 & 0.762 & 0.775 & 0.570 & 0.068 & 0.132 & 0.466 & 0.000 & 0.388 \\
    \muvit$_{[1,8,32]}$ {\small +Mask2Former} ({\small naive}) & $3 \times 256 \times 256$  & 0.924 & 0.851 & 0.791 & 0.764 & 0.852 & 0.930 & 0.849 & 0.644 & 0.756 & 0.831 & 0.832 & 0.820 \\
    \muvit$_{[1,8,32]}$ {\small +Mask2Former} & $3 \times 256 \times 256$ & \textbf{0.956} & \textbf{0.934} & \textbf{0.896} & \textbf{0.859} & 0.922 & \textbf{0.966} & \textbf{0.932} & \textbf{0.873} & \textbf{0.819} & \textbf{0.859} & \textbf{0.891} & \textbf{0.901} \\
    \muvit$_{[1,8,32]}$ {\small +UNETR} ({\small naive}) & $3 \times 256 \times 256$  & 0.919 & 0.837 & 0.773 & 0.718 & 0.860 & 0.918 & 0.833 & 0.556 & 0.734 & 0.808 & 0.767 & 0.793 \\
    \muvit$_{[1,8,32]}$ {\small +UNETR} & $3 \times 256 \times 256$  & 0.954 & 0.928 & 0.889 & 0.796 & \textbf{0.924} & 0.958 & 0.922 & 0.802 & 0.809 & 0.842 & 0.887 & 0.883 \\
    \bottomrule
    \end{tabular}
    }%
    \label{tab:brain}
    \end{table*}

The best-performing baseline, DeepLabV3 ($mDSC=0.843$), requires a substantially larger input size ($\sizetwo{1024}{1024}$) to achieve such performance, and drops significantly when using input sizes of $\sizetwo{512}{512}$ and $\sizetwo{256}{256}$ ($mDSC=0.676 \text{ and } 0.520 $ respectively). In contrast \muvit achieves superior results to the baselines with much smaller patches ($\sizethree{3}{256}{256}$) allowing the global anatomical context to be used through coarser resolution levels without requiring to increase the spatial dimensions of the input (and thus the memory footprint). An example of one of the harder areas to identify is the \textit{Septal complex},  a small region whose identification depends heavily on global anatomical positioning: \muvit$_{[1,8,32]}$ reaches a Dice of 0.859 compared to 0.734 for DeepLabV3, while most single-scale methods fail to classify it reliably.
\Cref{fig:seg_mouse} illustrates the general qualitative improvement brought by \muvit, showing that our multi-resolution approach correctly identifies precise anatomical boundaries that are missed by single-resolution baselines.

\subsection{Semantic segmentation on \datakidney}

\begin{figure*}[t]
    \centering
    \includegraphics[width=\linewidth]{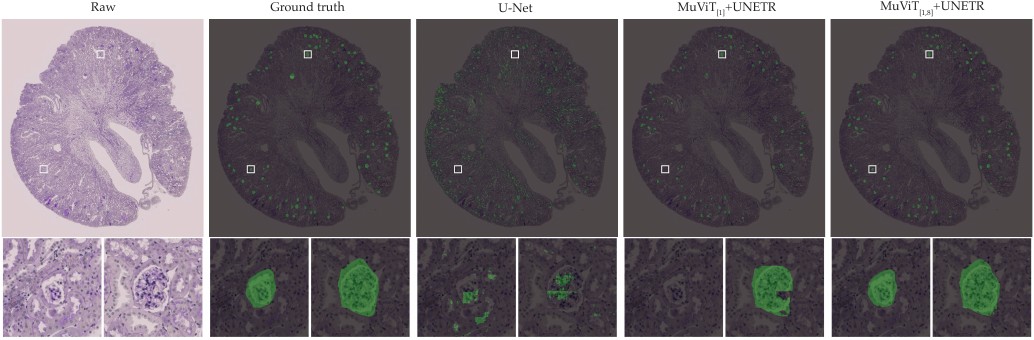}
    \caption{Semantic segmentation results on \datakidney. Depicted in different columns are the input image, ground truth, U-Net, \muvit$_{[1]}$+UNETR, and \muvit$_{[1,8]}$+UNETR predictions. Top: full image ($\sizetwo{52k}{56k}$ px); bottom: magnified regions (white boxes) showing detail comparison.}
     \label{fig:seg_kpis}
  \end{figure*}

\begin{table}[b]
  \caption{Binary semantic segmentation results on \datakidney{} (test set). * indicates the score was taken from \cite{tang2024}. All baselines operate at the highest resolution (corresponding to $l=1$).}
  \label{tab:kpis}
  \centering
  \begin{tabular}{@{}lrr}
    \toprule
    Method & Input size & DSC \\
    \midrule
    nnU-Net$^*$ \cite{isensee2021} & $\sizetwo{1024}{1024}$ & 0.6219 \\
    UNETR$^*$ \cite{hatamizadeh2022} & $\sizetwo{1024}{1024}$ & 0.6425 \\
    SegFormer$^*$ \cite{xie2021segformer} & $\sizetwo{1024}{1024}$ & 0.6501 \\
    DeepLabV3$^*$ \cite{chen2018encoder} & $\sizetwo{1024}{1024}$ & 0.6822 \\
    SwinUNETR-V2$^*$ \cite{he2023swinunetr} & $\sizetwo{1024}{1024}$ & 0.7158 \\
    SAM-ViT-B$^*$ \cite{kirillov2023segment} & $\sizetwo{1024}{1024}$ & 0.7519 \\
    SAM-ViT-H$^*$ \cite{kirillov2023segment} & $\sizetwo{1024}{1024}$ & 0.7724 \\
    HoloHisto-4K$^*$ \cite{tang2024} & $\sizetwo{3840}{2160}$ & 0.8454 \\
    \midrule
    U-Net \cite{ronneberger2015} & $\sizetwo{512}{512}$ & 0.5499\\
    \muvit{}$_{[1]}${\small +Mask2Former} & $\sizetwo{512}{512}$ & 0.8390\\
    \muvit{}$_{[1]}${\small +UNETR} & $\sizetwo{512}{512}$ & 0.8492\\
    \muvit{}$_{[1,8]}${\small +Mask2Former} & $\sizethree{2}{512}{512}$ & 0.8866\\
    \muvit{}$_{[1,8]}${\small +UNETR} & $\sizethree{2}{512}{512}$ & \textbf{0.8958}\\
    \bottomrule
  \end{tabular}
\end{table}
 
We finally evaluate \muvit{} on the kidney pathology segmentation task, comparing against both classical architectures and recent transformer-based methods as reported in~\cite{tang2024}. Similarly to the previous dataset, we obtain predictions for the whole WSIs through tiled inference without any overlap between tiles. \Cref{tab:kpis} shows the Dice score on the test set. \muvit{}$_{[1,8]}$+UNETR achieves a Dice of 0.8958, substantially outperforming HoloHisto-4K (0.8454), a method tailored for \datakidney, despite using much smaller inputs ($\sizethree{2}{512}{512}$ vs. $\sizetwo{3840}{2160}$). Notably, $\muvit_{[1,8]}$+UNETR also improves markedly over its single-resolution variant \muvit{}$_{[1]}$+UNETR (Dice 0.8958 vs. 0.8492), demonstrating the benefits of incorporating contextual information from coarser scales. Qualitative results in \Cref{fig:seg_kpis} show that \muvit{} produces sharper boundaries and captures fine tissue structures more reliably than U-Net and single-level \muvit$_{[1]}$ baselines, particularly in the magnified regions.

\begin{table}[t!]
  \caption{Linear probing results on \datakidney{} test set (4 classes) for HIPT~\cite{chen_scaling_2022} and \muvit encoders operating at different resolution sets. Reported is one-vs.-all (OvA) averaged ROC-AUC. Note that HIPT was used off-the-shelf and thus was not pre-trained on the \datakidney dataset.}
  \label{tab:kpis_probing}
  \centering
  \begin{tabular}{@{}lccccc@{}}
    \toprule
    Method  & FOV size & ROC-AUC\\
    \midrule
    HIPT ~\cite{chen_scaling_2022} (off-the-shelf) & \sizetwo{4096}{4096} & 0.907\\
    \muvit{}$_{[1]}$ & \sizetwo{256}{256} & 0.958 \\
    \muvit{}$_{[1,8]}$ & \sizethree{2}{256}{256} & 0.963 \\
    \muvit{}$_{[1,8,32,64]}$ & \sizethree{4}{256}{256} & \textbf{0.988}\\
    \bottomrule
  \end{tabular}
\end{table}

To further analyze the learned representations, we perform a linear probing experiment using features from a frozen MAE-pretrained encoder on 4 resolution levels ($1, 8, 32, 64$). We  train a logistic regression model on top of the extracted features in order to classify the experimental group of each WSI (4 subgroups). For this, we create an offline training set by sampling five random nested multi-resolution crops (see \Cref{sec:methods}) per training image and a corresponding test set using 10 crops per test image. \Cref{tab:kpis_probing} shows the one-vs-all (OVA) ROC-AUC scores for the 4-class classification problem demonstrating the effectiveness of \muvit for capturing information across scales: adding more resolution levels progressively improves performance from 0.958 (single-level) to 0.988 (four levels $[1,8,32,64]$). This indicates that the encoder learns increasingly rich representations as more scales are incorporated. We also include a comparison on embeddings obtained  using HIPT~\cite{chen_scaling_2022} off-the-shelf (ROC-AUC 0.907 with input size \sizetwo{4096}{4096}). Note that HIPT was pre-trained on a large pool of histopathology datasets which does not include \datakidney.

\subsection{Coordinate sensitivity and scaling behaviour}

Despite world coordinate alignment across multiple scales being essential for performance as highlighted by the \textit{naive} experiments (\cref{tab:synthetic,tab:brain,tab:kpis}), we wanted to test whether \muvit models exhibit robustness to coordinate noise during inference. Note that we include an augmentation which adds scaling and shift to the samples bounding box coordinates during pre-training and fine-tuning.

We thus performed such an analysis by measuring the segmentation performance of the best performing \muvit models for each dataset while adding Gaussian noise with standard deviation $\sigma$ to the inference bounding box coordinates. We observe that \muvit remains robust to such noise, with only minor degradation up to $\approx 32 \text{px}$ on all datasets (\Cref{fig:robustness}).

We also benchmarked runtime and peak memory on arbitrarily-sized inputs using a fixed patch size $p=8$, as used throughout the paper, as well as an increasing number of resolution levels $l \in \{1,8,32\}$. \Cref{fig:overhead} expectedly showcases that \muvit incurs computational overhead in exchange for a substantially larger effective context.
\section{Conclusions}
\label{sec:conclusions}

We presented \muvit, a multi-resolution Vision Transformer architecture for large-scale microscopy image analysis. The architecture processes multiple resolution levels of the same spatial region through a unified encoder, using coordinate-based Rotary Position Embeddings to preserve spatial relationships across scales. This explicit geometric encoding enables cross-scale attention between true multi-resolution observations, which differs from hierarchical architectures that rely on learned feature pyramids.
Our experiments demonstrate that multi-resolution processing with accurate spatial information provides substantial benefits across synthetic and real microscopy datasets. The \textit{naive bbox} baseline confirms that accurate spatial correspondence is crucial for performance, despite \muvit exhibiting robustness when injecting noise into the world coordinates. Linear probing experiments show progressive improvement as more resolution levels are added, indicating that the encoder learns increasingly rich representations with explicit multi-scale structure. While jointly attending over all scales increases computational cost, this overhead could be reduced in future work through sparse or cross-scale attention schemes.

We have focused on semantic segmentation as a downstream task as it most directly probes the core challenge addressed by \muvit: combining global (\eg anatomical) context with local, high resolution detail in gigapixel images. Assessing the impact on other common tasks for microscopy image processing, such as instance segmentation or object detection, is a natural direction for future work.

The framework is flexible and can incorporate non-nested views from different spatial regions of an image, and can be readily extended to 3D volumes where crops can be sampled from different slices while maintaining spatial consistency through coordinate-based positional encoding.

\section*{Author contributions}

A.D.M. developed the idea, implemented the software, conducted experiments and interpreted the results. G.L.M. contributed to the conceptual development, helped shape the experimental design and interpreted results. M.W. conceived the project, developed the idea, designed the architecture, implemented the software, conducted experiments and interpreted the results. G.L.M. and M.W. jointly supervised the work. All authors discussed the results and wrote the manuscript.

\section*{Acknowledgements}

We thank members of the Weigert and La Manno labs for their feedback and discussions of the project. G.L.M. received support from the Swiss National Science Foundation grant PZ00P3\_193445 and from the Horizon Europe ERC Starting grant 101221782. A.D.M. and G.L.M. were supported by the Carl Zeiss AG-EPFL joint initiative Research-IDEAS. M.W. was supported by the BMFTR and by SMWK (project identification number: ScaDS.AI). 
{
    \small
    \bibliographystyle{ieeenat_fullname}

}

\clearpage
\setcounter{page}{1}
\maketitlesupplementary
\renewcommand{\thesection}{\Alph{section}}
\renewcommand{\thesubsection}{\thesection.\arabic{subsection}}
\setcounter{section}{0}
\setcounter{figure}{0}
\setcounter{table}{0}
\renewcommand{\thefigure}{S\arabic{figure}}
\renewcommand{\thetable}{S\arabic{table}}

\section{Training details and hyperparameters}
\label{sec:suppl_training}

This section provides comprehensive implementation details for reproducibility. Note that the full codebase (including experiments code) and datasets will be released upon publication.

\subsection{Hardware specifications}

All models were trained on the \textit{Capella} HPC cluster at TU Dresden. Its nodes consist of 2x AMD EPYC 7282 CPUs (64 cores). Runs booked up to 256 GB RAM (DDR-4800). For MAE pre-training, models were trained on a single node in a multi-GPU setting (4x Nvidia H100, 96GB). All semantic segmentation models (including the baselines) were trained on a single node using a single GPU (1x Nvidia H100, 96 GB)

\subsection{MAE pre-training}

\textbf{Architecture.} The encoder consists of 12 transformer layers with embedding dimension $D=512$ and patch size $P=8$. Each resolution level has independent patch embedding layers followed by learnable level embeddings. The decoder uses $L$ lightweight networks (one per level), each with 2 transformer layers and dimension 256. The first decoder layer incorporates cross-attention to the full set of visible encoder outputs with coordinate-based RoPE. All transformer layers (including those in the decoder) use the coordinate-based RoPE embeddings described in the main text.

\textbf{Masking strategy.} We use a masking ratio of $\rho=0.75$ with Dirichlet-weighted sampling across resolution levels. The proportions of masked tokens per level $(\lambda_1, \ldots, \lambda_L)$ are sampled from $\text{Dir}(\alpha)$ with $\alpha=0.5$, encouraging the model to learn cross-scale relationships by varying the relative masking ratio per level.

\textbf{Data sampling.} Due to the large-scale nature of the data, we employ a sampling-based strategy in which a nested multi-resolution crop is randomly sampled from each image (see \Cref{sec:experiments}). We therefore use the term \textit{training sample} to denote such a multi-resolution crop extracted from a training image. We define an \textit{epoch} as a pass of 10000 training samples to the network during training. Sampled crops are normalized with percentile-based normalization ($p_m, p_M = 1, 99.8$) for \datamouse and by dividing by 255 for \datakidney.

\textbf{Multi-resolution configuration.} For \datakidney, we use resolution levels $[1, 8, 32, 64]$ during MAE pre-training. For \datamouse, we use levels $[1, 8, 32]$ for both MAE pre-training. All levels use the same pixel-space patch size (\sizetwo{256}{256} pixels), with lower resolution levels capturing progressively larger spatial context.

\textbf{Optimization.} MAE pre-training uses a batch size of 48/64 for 2000/1000 epochs for \datamouse/\datakidney epochs with learning rate $6\times10^{-5}$. We use the AdamW optimizer with gradient clipping at 0.5 and deterministic cosine annealing learning rate schedule. Early stopping is not used.

\textbf{Loss function.} The loss function for MAE pre-training is the mean squared error (MSE) (computed on masked patches only) averaged across all resolution levels.

\textbf{Data augmentation.} Used augmentations include random horizontal and vertical flips as well as intensity jittering and bounding box perturbations (random shifts and scaling).  Note that the horizontal and vertical flips affect the whole nested crop (\ie they are not applied independently to each resolution level) and that bounding boxes are augmented accordingly.

\subsection{Segmentation fine-tuning}

\textbf{Architecture.} We use the same pre-trained encoder architecture (12 transformer layers, token embedding dimension $D=512$) and couple it with either a UNETR-style decoder or Mask2Former decoder as described in \Cref{sec:methods}.

\textbf{Data sampling.} We use the same sampling strategy as for MAE pre-training. Note that the target semantic segmentation mask is only sampled at the highest 
resolution level. For \datakidney we ensure that at least half the training samples contain at least a positive class pixel. We define an \textit{epoch} as a pass of 4096 training samples to the network during training for the semantic segmentation task. Sampled crops are normalized with percentile-based normalization ($p_m, p_M = 1, 99.8$) for \datamouse and by dividing by 255 for \datakidney.

\textbf{Multi-resolution configuration.} For \datakidney, we extract only levels $[1, 8]$ from the MAE pretrained encoder for segmentation fine-tuning (as shown in \cref{tab:kpis}). For \datamouse, we use the full encoder. All levels use the same pixel-space patch size (\sizetwo{256}{256} pixels for \datamouse, \sizetwo{512}{512} for \datakidney differing from the pre-training patch size).

\textbf{Optimization.} Semantic segmentation training uses a batch size of 16 for \datamouse (except \muvit{\small{+UNETR}} variants which used a batch size of 8 due to memory limitations) and a batch size of 8 for \datakidney. All our models were trained for 100 epochs. We employ asymmetric learning rates: decoder learning rate $1\times10^{-4}$, encoder learning rate $1\times10^{-5}$ (0.1$\times$ decoder rate). We use the AdamW optimizer with gradient clipping at 0.5 and cosine annealing schedule.

\textbf{Loss function.} We use a combined cross-entropy and Dice loss: $\mathcal{L} = \lambda_{\text{CE}} \cdot \mathcal{L}_{\text{CE}}(\tilde{y}, y) + \lambda_{\text{Dice}} \cdot \mathcal{L}_{\text{Dice}}(\tilde{y}, y)$ with $\lambda_{\text{CE}}=1.0$ and $\lambda_{\text{Dice}}=0.1$. For \datakidney, we double the weight to the foreground class to aid with the imbalance between foreground and background. The Dice loss excludes the background class (class 0) and is computed separately per resolution level before averaging.

\textbf{Data augmentation.} We use the same augmentation set as in the MAE pre-training case.

\section{\muvit components}

\subsection{Number of parameters}

\begin{table}[htbp!]
  \caption{Number of parameters for \muvit components and variants.}
  \label{tab:muvit_complexity}
  \centering
  \resizebox{\columnwidth}{!}{%
  \begin{tabular}{@{}lc@{}}
    \toprule
    Module & Parameters \\
    \midrule
    \muvit{}$_{[1]}$ {\small{(encoder)}} & 25.24M \\
    \muvit{}$_{[1,8,32]}$ {\small{(encoder)}} & 25.31M \\
    \midrule
    \muvit{}$_{[1]}${\small{+MAE decoder}} & 26.44M \\
    \muvit{}$_{[1,8,32]}${\small{+MAE decoder}} & 28.88M \\
    \midrule
    \muvit{}$_{[1]}${\small{+Mask2Former decoder}} & 33.44M \\
    \muvit{}$_{[1,8,32]}${\small{+Mask2Former decoder}} & 36.80M \\
    \muvit{}$_{[1]}${\small{+UNETR decoder}} & 31.66M \\
    \muvit{}$_{[1,8,32]}${\small{+UNETR decoder}} & 35.02M \\
    \bottomrule
  \end{tabular}%
  }
\end{table} 

\subsection{Semantic Segmentation Decoder}

\paragraph{UNETR decoder}
The \emph{UNETR} decoder uses skip connections from multiple intermediate encoder layers to preserve fine-grained spatial information during upsampling, as introduced in the original UNETR work. The decoder consists of a progressive upsampling pathway that combines transformer-encoded features with skip connections at multiple resolution levels. We extract intermediate features from arbitrarily selected encoder layers (by default, evenly spaced across the encoder depth) and project them to a $D$-dimensional vector ($D=256$). The decoder operates through multiple upsampling stages, where each stage performs bilinear upsampling followed by residual convolutional blocks that fuse the upsampled features with corresponding skip connections from the intermediate encoder outputs. The last skip connection involves directly processing the original input image at the highest resolution level instead of the encoder outputs.
Each upsampling stage consists of bilinear interpolation followed by two regular convolutional blocks (\sizetwo{3}{3} convolutional layers with batch normalization and a ReLU activation). Skip connections are incorporated through element-wise addition after using \sizetwo{1}{1} convolutions to match the channel axes. The final output layer produces segmentation predictions at full input resolution through a \sizetwo{1}{1} convolution that maps from the embedding dimension $D$ to the target number of output channels/classes.

\paragraph{Mask2Former decoder}
The \emph{Mask2Former} decoder follows the query-based segmentation paradigm introduced in the original work. The decoder consists of three main components: a pixel decoder that progressively upsamples encoder features to full resolution, a set of learnable mask queries that represent potential objects, and a transformer decoder that refines these queries. For the pixel decoder we use a series of bilinear upsampling layers followed by regular convolutional blocks (as described above) in order to create dense mask features at the original image resolution. We initialize $|M_q|$ learnable embeddings of dimension $D=256$ that serve as mask queries, where the number of queries is set to be at least equal to the number of output classes.
The transformer decoder refines the mask queries using multiple transformer layers. Each layer applies self-attention among the queries to model relationships between different potential masks, followed by cross-attention between queries and the encoder features. We use RoPE for the spatial features obtained from the encoder to maintain spatial awareness during cross-attention, while queries remain position-agnostic so that they can be used as learnable representations. We then feed the refined queries to a layer normalization module followed by an MLP to produce the mask embeddings.
The final segmentation masks are generated by computing the dot product between these mask embeddings and the pixel decoder's mask features, followed by a refinement convolution that maps from $|M_q|$ to the desired number of output channels.

\clearpage

\section{Semantic segmentation results}
\label{sec:suppl_semantic}
\subsection{\datamouse}

{\raggedright
\noindent\begin{minipage}{\textwidth}
    \includegraphics[width=\textwidth]{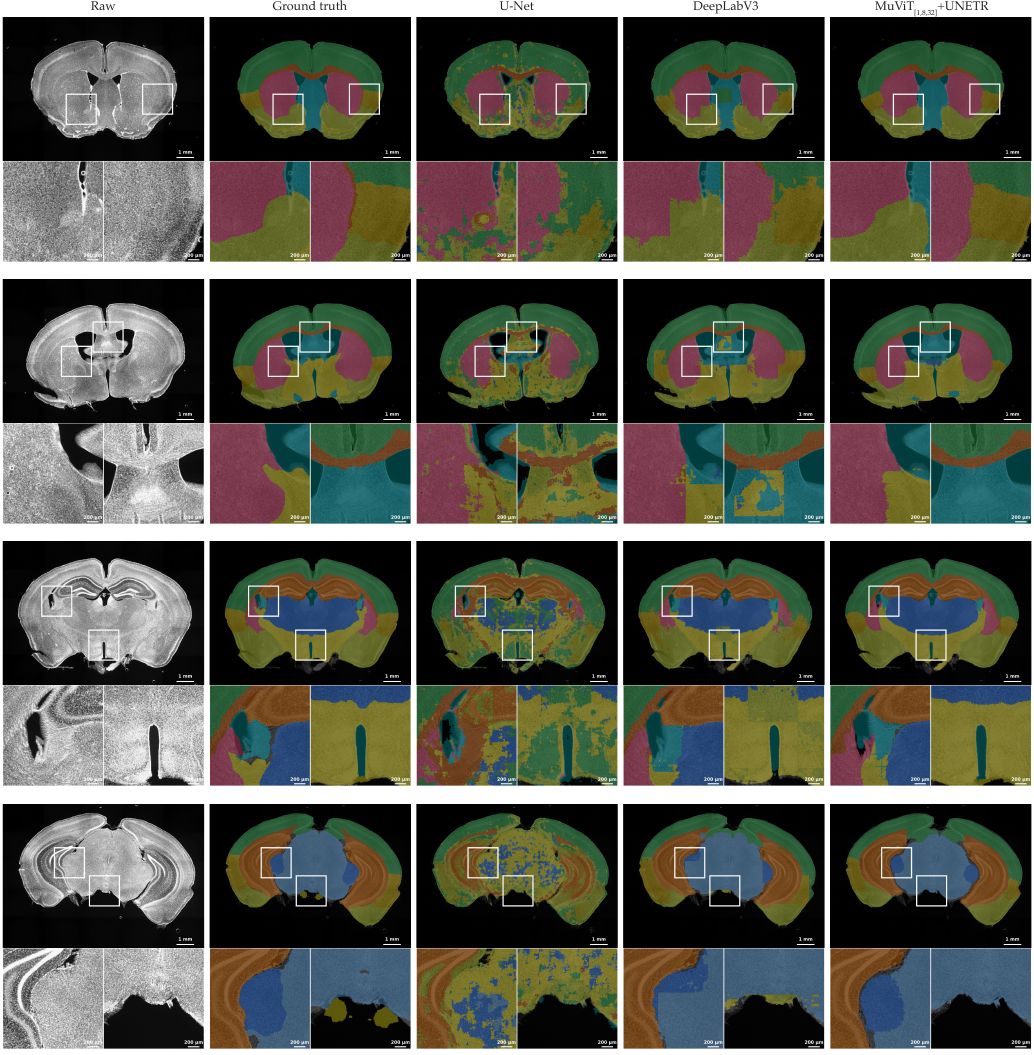}
    \captionof{figure}{Semantic segmentation results on four test images of \datamouse. The images correspond to different brain sections of the same specimen, exhibiting anatomical changes and, thus, showcasing different semantic classes predictions. Columns show (left to right): input image, ground truth, U-Net, DeepLabV3, and \muvit{}$_{[1,8,32]}$+UNETR semantic predictions. The input sizes of the baselines are $\sizetwo{1024}{1024}$ while the input for \muvit{}$_{[1,8,32]}$+UNETR is $\sizethree{3}{256}{256}$. Magnified regions (white boxes) show detailed comparison.}
     \label{fig:seg_mouse_suppl}
\end{minipage}
\par}

\clearpage

\subsection{\datakidney}

{\raggedright
\noindent\begin{minipage}{\textwidth}
    \includegraphics[width=\textwidth]{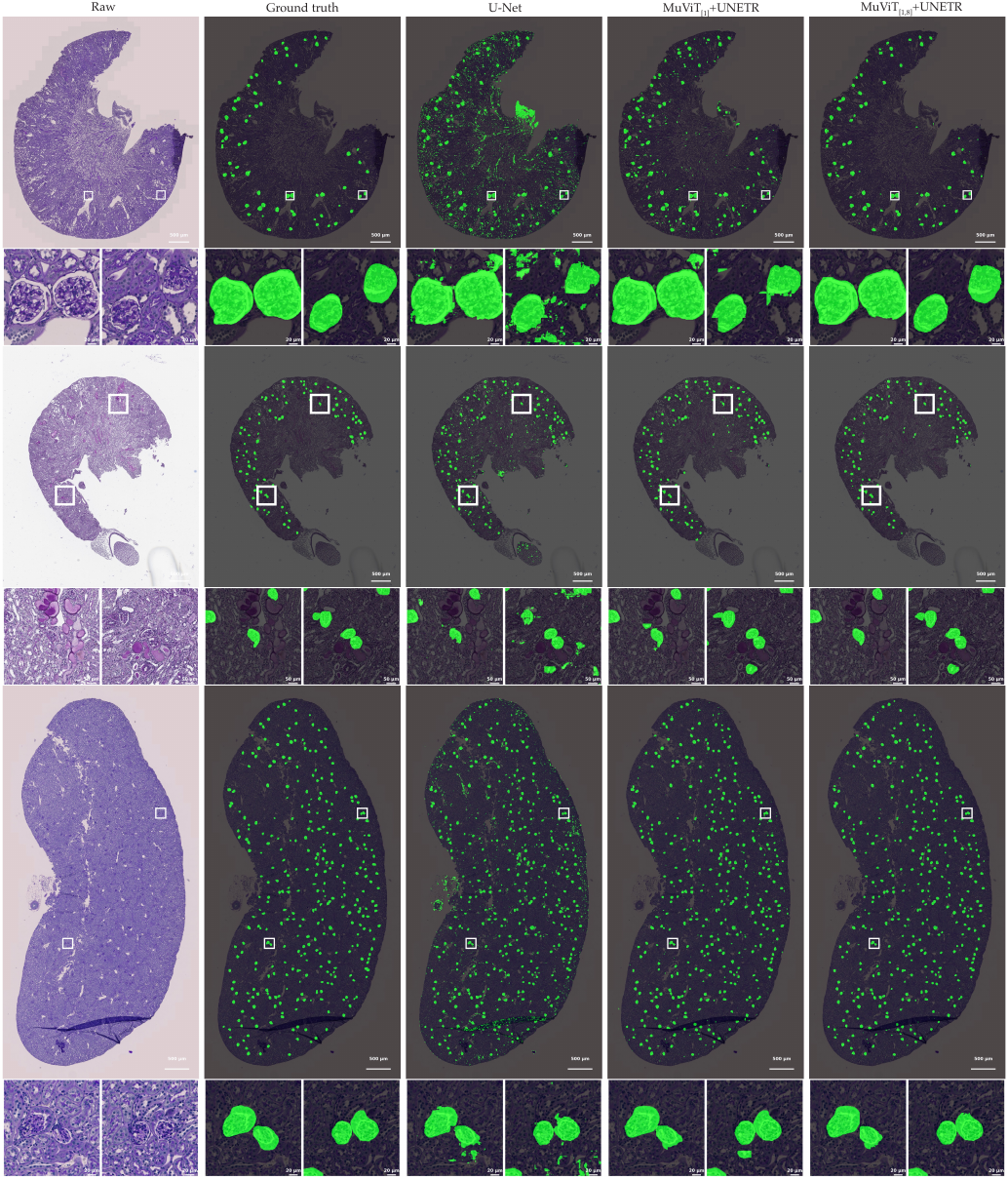}
    \captionof{figure}{Semantic segmentation results on three test whole slide images (WSIs) of \datakidney. The images correspond to three different conditions. Columns show (left to right): input image, ground truth, U-Net, \muvit{}$_{[1,8]}$+UNETR, and \muvit{}$_{[1,8]}$+UNETR semantic predictions. The spatial input size of all models is $\sizetwo{512}{512}$. Magnified regions (white boxes) show detailed comparison.}
     \label{fig:seg_kidney_suppl}
\end{minipage}
\par}

\clearpage

\begin{figure*}[t]
\subsection{Coordinate sensitivity and scaling behaviour}
  \centering
\begin{minipage}[t]{0.3\linewidth}
       \centering
       \includegraphics[width=0.835\linewidth]{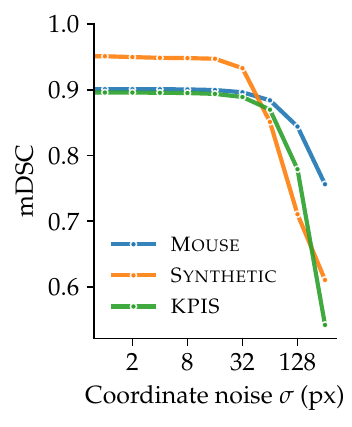}
       \caption{Effect of adding Gaussian noise of varying standard deviation $\sigma$ on \muvit segmentation performance across different datasets.}
       \label{fig:robustness}
\end{minipage}\quad\quad
\begin{minipage}[t]{0.6\linewidth}
       \centering
       \includegraphics[width=\linewidth]{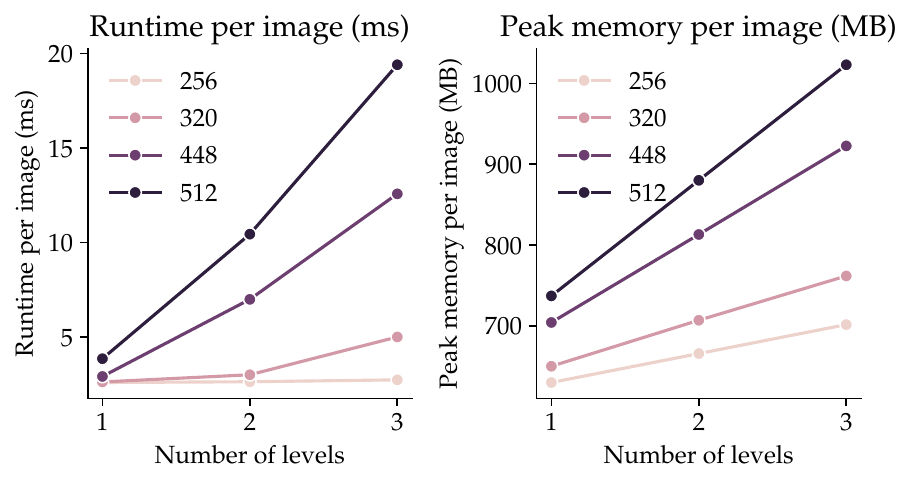}
       \caption{Runtime and memory scaling of \muvit. Hue denotes input spatial dimensions.}
       \label{fig:overhead}
\end{minipage}
\end{figure*}

\vspace*{\fill}
 
\end{document}